\documentclass[sigconf]{acmart}
\usepackage{graphicx,subcaption,caption}
\usepackage{multirow}
\usepackage{amsmath}
\usepackage{makecell}
\usepackage{utfsym}
\usepackage{pifont}
\usepackage{colortbl}
\usepackage{color, xcolor} 
\usepackage{tcolorbox}
\usepackage{bbm}
\usepackage{dsfont} 
\tcbuselibrary{most}
\usepackage{listings}
\usepackage{tabularx}
\usepackage{array}
\usepackage{booktabs}
\usepackage[utf8]{inputenc}
\usepackage[T1]{fontenc}
\definecolor{HTML}{HTML}{E6E6E6}
\definecolor{HTML}{HTML}{FFD6E7}
\definecolor{mycolor1}{RGB}{206, 220, 233} 
\definecolor{mycolor2}{RGB}{79, 123, 168} 
\definecolor{mycolor3}{RGB}{247, 203, 204} 
\definecolor{mycolor4}{RGB}{235, 131, 134} 
\tcbset{
    appendixbox/.style={
        enhanced,
        colback=gray!10!white,
        colframe=black,
        fonttitle=\bfseries\small,
        fontupper=\small,
        boxsep=3pt,
        left=4pt,
        right=4pt,
        top=4pt,
        bottom=4pt,
        before skip=6pt,
        after skip=6pt,
        breakable=true 
    }
}

\newtcolorbox[auto counter, number within=section]{numberedappendixbox}[2][]{
    appendixbox,
    title={\thetcbcounter. #2},
    #1
}
\usepackage{hyperref}
\AtBeginDocument{%
  }

\setcopyright{acmlicensed}
\copyrightyear{2018}
\acmYear{2018}
\acmDOI{XXXXXXX.XXXXXXX}
\acmConference[xx]{Make sure to enter the correct
  conference title from your rights confirmation email}{xx}{xx}
\acmISBN{978-1-4503-XXXX-X/2018/06}

\begin{document}

\title{Towards Real-Time Fake News Detection under Evidence Scarcity}

\author{Guangyu Wei}
\authornote{Equal contribution.}
\affiliation{
  \institution{Nanjing University$^{\ddag}$}
  \authornote{Work done while interning at Nanjing University.}
  \institution{Ocean University of China}
    \city{Suzhou}
  \country{China}}
\email{wgy3129@stu.ouc.edu.cn}

\author{Ke Han}
\authornotemark[1]
\affiliation{%
  \institution{University of Trento}
  \city{Trento}
  \country{Italy}}
  \email{ke.han@unitn.it}

\author{Yueming Lyu}
\authornote{Corresponding Author}
\affiliation{%
  \institution{Nanjing University}
    \city{Suzhou}
  \country{China}}
\email{ymlv@nju.edu.cn}

\author{Yu Luo}
\affiliation{%
    \institution{Ocean University of China}
      \city{Qingdao}
    \country{China}}
  \email{luoyu@stu.ouc.edu.cn}

\author{Yue Jiang}
\affiliation{%
 \institution{Chinese Academy of Sciences}
  \city{Beijing}
 \country{China}}
 \email{yue.jiang@cripac.ia.ac.cn}

\author{Caifeng Shan}
\affiliation{%
  \institution{Nanjing University}
    \city{Suzhou}
  \country{China}}
  \email{cfshan@nju.edu.cn}

\author{Nicu Sebe}
\affiliation{%
  \institution{University of Trento}
    \city{Trento}
  \country{Italy}}
  \email{niculae.sebe@unitn.it}

\renewcommand{\shortauthors}{Guangyu Wei et al.}


\begin{abstract}
Fake news detection becomes particularly challenging in real-time scenarios, where emerging events often lack sufficient supporting evidence. Existing approaches often rely heavily on external evidence and therefore struggle to generalize under evidence scarcity.
To address this issue, we propose Evaluation-Aware Selection of Experts (EASE), a novel framework for real-time fake news detection that dynamically adapts its decision-making process according to the assessed sufficiency of available evidence.
EASE introduces a sequential evaluation mechanism comprising three independent perspectives:
(1) Evidence-based evaluation, which assesses evidence and incorporates it into decision-making only when the evidence is sufficiently supportive;
(2) Reasoning-based evaluation, which leverages the world knowledge of large language models (LLMs) and applies them only when their reliability is adequately established; and
(3) Sentiment-based fallback, which integrates sentiment cues when neither evidence nor reasoning is reliable.
To enhance the accuracy of evaluation processes, EASE employs instruction tuning with pseudo labels to guide each evaluator in justifying its perspective-specific knowledge through interpretable reasoning.
Furthermore, the expert modules integrate the evaluators’ justified assessments with the news content to enable evaluation-aware decision-making, thereby enhancing overall detection accuracy.
Moreover, we introduce RealTimeNews-25, a new benchmark comprising recent news for evaluating model generalization on emerging news with limited evidence. Extensive experiments demonstrate that EASE not only achieves state-of-the-art performance across multiple benchmarks, but also significantly improves generalization to real-time news.
The code and dataset are available: https://github.com/wgyhhhh/EASE.

\end{abstract}

\vspace{-2mm}




\maketitle

\section{Introduction}


With the rapid growth of web-based media such as online news platforms and social networks, misinformation spreads at unprecedented speed and scale.
The circulation of fake news on the Web poses significant risks to politics~\citep{3}, economy~\citep{2}, and public safety~\citep{1}.
Therefore, developing automated fake news detection methods is essential for maintaining information integrity and trustworthiness on the Web, making this research highly relevant to the Web community.

\begin{figure}[t]
    \centering
    \includegraphics[width=\linewidth,height=0.4\textwidth]{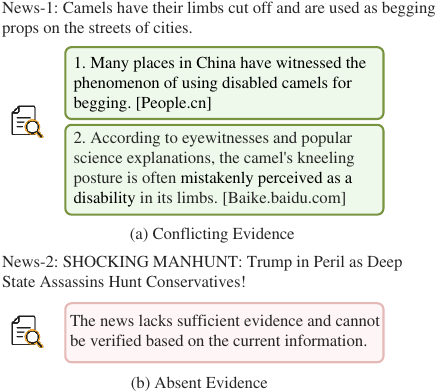}
    \caption{Examples of limited supporting evidence: (a) conflicting information across different sources, and (b) the absence of relevant evidence.}
    \label{fig_intro}
\end{figure}

Traditional fake news detection approaches \citep{4,5,11,12} typically rely solely on news content for decision-making, which often limits their generalizability to unseen news due to out-of-distribution issues.
To overcome this limitation, recent studies have explored retrieval-augmented methods that retrieve external evidence from web-scale or structured repositories to support decision-making \citep{7,8,9,10,14}.

However, real-time news, where timeliness is a defining characteristic, typically involves emerging or rapidly evolving events. In such scenarios, news items are rarely fact-checked immediately, as supporting evidence is often scarce, unreliable, or entirely unavailable. This lack of evidence makes evidence-based approaches struggle to produce accurate predictions, thereby limiting their applicability to real-time detection settings.
To address this challenge, this paper investigates \textbf{real-time fake news detection with limited supporting evidence}, aiming to enhance detection effectiveness in real-world, time-sensitive contexts.

Specifically, we systematically address three key challenges in this largely underexplored area:

1) Evaluating the sufficiency of retrieved evidence. 
While previous studies often assess evidence credibility based primarily on source reliability \citep{60,62,63,24}, emerging events frequently lack verified information from authoritative outlets, limiting the effectiveness of such approaches.
Furthermore, evidence collected from unofficial or unverified sources is often informal, conflicting, or even absent (as illustrated in Fig.~\ref{fig_intro} (1) and (2)), which may cause models to draw unreliable conclusions.

2) Handling insufficient or unreliable evidence.
When evidential support is lacking, designing an effective fallback mechanism that allows models to assess news authenticity from alternative perspectives remains a critical yet unresolved challenge.

3) Evaluating model generalization under evidence scarcity. 
Existing public benchmarks composed primarily of historical news \citep{57,58,59}, where abundant evidence or prior knowledge is available to verify authenticity. However, real-time fake news detection requires evaluation on recent and dynamically evolving news data to more accurately assess model generalization under the practical challenge of evidence scarcity.

To address these challenges, we propose \textbf{Evaluation-Aware Selection of Experts (EASE)}, a novel framework for real-time fake news detection.
EASE introduces a sequential evaluation mechanism that assesses the quality of available evidence and adaptively selects the most appropriate decision-making strategy for each news instance.
Specifically, it dynamically incorporates three independent perspectives:
1) \textbf{Evidence-based}: An evaluator performs large-scale online retrieval of external evidence and thoroughly analyzes their usability. If the retrieved evidence is deemed sufficiently supportive, an evidence expert leverages it, along with the evaluator’s justification, to make authenticity predictions.
2) \textbf{Reasoning-based}: When external evidence is insufficient, EASE activates the internal reasoning capabilities of large language models (LLMs) to infer conclusions using world knowledge. A reasoning evaluator examines the reliability of the inferred logic before authorizing a reasoning expert to make a final decision.
3) \textbf{Sentiment-based}: If neither evidence nor reasoning provides reliable verification, a sentiment expert serve as a fallback strategy to examine emotional tone, subjectivity, and stylistic cues for authenticity assessment.

More importantly, both the evaluator and expert modules are carefully designed to ensure reliability and interpretability.
Instead of directly employing LLMs as evaluators, 
we propose an instruction-tuning strategy coupled with pseudo-label supervision to fine-tune LLMs.
This process provides feedback to the model’s outputs, guiding it to rigorously justify the reliability of perspective-specific knowledge and to generate faithful, interpretable reasoning rather than speculative explanations.
Meanwhile, the expert modules are designed not to rely solely on available knowledge for decision-making but to integrate the justified evaluations from the evaluators with their interactions with the news content, thereby enhancing overall decision accuracy and robustness.

Moreover, to advance research on real-time fake news detection, we introduce a new benchmark, \textbf{RealTimeNews-25}, consisting of 3,487 news articles collected between June 2024 and September 2025.
The dataset covers recent and rapidly evolving events characterized by limited supporting evidence, providing a challenging and timely benchmark for evaluating model robustness in real-world, time-sensitive scenarios.

We conduct extensive experiments on RealTimeNews-25 and the widely used benchmarks Weibo \citep{57},  Weibo21 \citep{58} and GossipCop \citep{59}.
Experimental results show that our approach not only achieves state-of-the-art accuracy on historical news but also substantially improves generalization to real-time news with limited evidence, highlighting its effectiveness in practical scenarios.
The main contributions of this paper are summarized as follows:

\begin{itemize}
\item We systematically investigate real-time fake news detection under evidence scarcity, a rarely explored yet highly practical problem. We propose EASE, a novel framework that dynamically evaluates evidence quality and adaptively selects the most trustworthy knowledge for decision-making.
\item We design an evaluator–expert training paradigm that fine-tunes LLMs via instruction tuning with pseudo labels, enabling evaluation-aware expert selection based on news characteristics and perspective-specific knowledge.
\item We construct RealTimeNews-25, a new benchmark dataset of emerging news events, to evaluate model generalization in real-world, time-sensitive scenarios.
\item Extensive experiments demonstrate that EASE achieves state-of-the-art performance on historical news across existing benchmarks, and significantly enhances generalization to real-time news with limited evidence.
\end{itemize}

\section{Related Work}

\noindent
\textbf{Fake News Detection.}
Existing approaches can generally be divided into two categories based on their use of external knowledge: content-based methods, which rely solely on news content, and knowledge-based methods, which incorporate additional information such as external evidence \citep{35,36}, user comments \citep{40,41}, or media background \citep{50}.

1) Content-based approaches.
These methods primarily analyze intrinsic features of news content, such as writing style, emotional tone, and image–text consistency.
For example, UEEI constructs an affective enhancement graph to model emotional conflicts between news and user comments \citep{30}.
FND-CLIP leverages a pre-trained CLIP model to guide multimodal information fusion \citep{51}, while MIMoE-FND introduces a mixture-of-experts mechanism that adaptively customizes fusion strategies for different scenarios \citep{5}.
However, these approaches heavily depend on specific training data, leading to limited generalization \citep{13} and weak interpretability.

2) Knowledge-based approaches.
These methods utilize external information to assist in verification.
For instance, GET proposes a unified evidence-graph-based framework for fake news detection \citep{52}, MUSER introduces a multi-step evidence retrieval framework that simulates human reasoning during news verification \citep{35}, and SEE integrates attention mechanisms with early termination to improve the use of unlabeled evidence \citep{53}.
ARG further incorporates LLM-generated rationales into a trainable detection framework \citep{37}.
Although these models capture dependencies between external evidence and news content, they are often hindered by evidence scarcity or unreliability, particularly in real-time news detection where timely and trustworthy information is lacking.

\begin{figure*}[t]
    \centering
    \includegraphics[width=\linewidth, trim=0 0 0 0mm, clip]{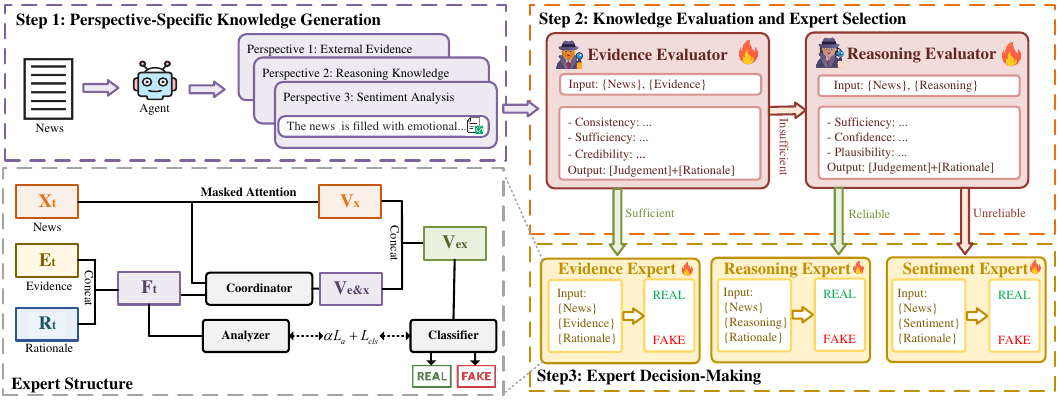}
    \vspace{-6mm}
    \caption{The Evaluation-Aware Selection of Experts (EASE) framework.
EASE first employs an evidence-based strategy, where (1) the evidence agent retrieves external evidence, (2) the evaluator assesses its sufficiency, and (3) the evidence expert makes the final decision. When the evidence is insufficient, EASE switches to a reasoning-based strategy that mirrors the evidence-based pipeline, generating and evaluating reasoning knowledge for decision-making. If the reasoning knowledge is unreliable, EASE activates a sentiment-based strategy, which analyzes stylistic and emotional cues to produce the final judgment.
    }
    \label{fig:framework}
\end{figure*}

\noindent
\textbf{Fact-Checking.}
Fact-checking focuses on verifying the factual accuracy of claims, typically by leveraging external knowledge.
For instance, Hiss decomposes complex claims into sub-claims and verifies them sequentially using retrieved evidence \citep{10}.
DEFAME \citep{14} and Veract Scan \citep{24} incorporate LLMs and web-based knowledge sources into verification pipelines to enhance performance.
However, their dependence on external knowledge suffers from insufficient or unreliable knowledge.
When knowledge is lacking, most fact-checking systems label a claim as \emph{unverifiable}, which limits their applicability to fake news detection, especially in real-time contexts where authoritative evidence is often unavailable.
For example, experiments on our RealTimeNews-25 dataset show that Hiss~\citep{10}, Veract Scan~\citep{24}, and DEFAME~\citep{14} output “Not Enough Information” labels for 49.6\%, 33.0\%, and 45.6\% of news samples, respectively, highlighting the challenge of evidence scarcity in real-time scenarios.

\noindent
\textbf{Evidence Credibility Evaluation.}
To address the issue of unreliable evidence, prior studies have explored credibility assessment mechanisms, primarily focused on evaluating the reliability of evidence sources.
For example, \citet{60} analyze media contexts to incorporate source credibility into fact-checking systems, while \citet{62,63,24} train classifiers using various features of media outlets to assess their trustworthiness.
However, in real-time news detection, emerging events often lack reliable sources due to their immediacy, making source-based evaluation less effective.
In contrast, our approach evaluates both evidence content and source reliability from multiple perspectives, such as consistency, sufficiency, and credibility, to achieve a more comprehensive assessment.
Moreover, EASE introduces fallback mechanisms to handle scenarios with insufficient or unreliable evidence.

\section{Methodology}

In this work, we propose Evaluation-Aware Selection of Experts (EASE), a novel framework for real-time fake news detection under evidence scarcity.

As illustrated in Fig.~\ref{fig:framework}, 
to address the potential scarcity or unreliability of evidence in real-time news, EASE introduces a sequential evaluation and expert selection pipeline from three perspectives:
1) Evidence-based decision: An evaluator first assesses the sufficiency of retrieved evidence. When the evidence is deemed sufficient, an expert leverages both the evidence and the evaluator’s justification for final decision-making.
2) Reasoning-based decision: If the evidence is insufficient, a reasoning evaluator examines the quality of internally generated reasoning knowledge from LLMs. When this reasoning is reliable, the corresponding expert is activated for decision-making.
3) Sentiment-based decision: When both evidence and reasoning are unreliable, a sentiment expert acts as a fallback, analyzing emotional tone and linguistic cues in the news content to support the final decision.



\subsection{Evidence-Based Decision-Making}
\label{3.1}

EASE prioritizes evidence-based decision-making for verifying news authenticity, as external evidence typically provides the most reliable basis for judgment. However, since such evidence is often insufficient or unreliable for emerging events and weakens its effectiveness, EASE introduces an evidence evaluator to assess whether the retrieved evidence is sufficiently supportive for decision-making.

\subsubsection{\textbf{Evidence Evaluator}}

\noindent
\textbf{(1) Evidence Agent.}
To gather external information, we employ an evidence agent that retrieves and summarizes web-based evidence.
Rather than relying on a single-pass search, the agent performs iterative retrieval guided by in-context prompting of LLMs~\citep{54}.
As illustrated in Figure~\ref{fig:agent_prompt}, the agent:
(1) generates search queries conditioned on the news content;
(2) leverages Google Search via the Serper API\footnote{\url{https://serper.dev}} to provide the top 3 relevant web pages matching the query;
(3) scrape the corresponding page using Firecrawl\footnote{\url{https://github.com/mendableai/firecrawl}}, then summarize and assess its relevance and reliability; and
(4) repeats the retrieval process when the collected evidence is deemed insufficient, up to a predefined iteration limit.
This iterative mechanism allows the agent to progressively refine the evidence set, ensuring both coverage and quality.

\noindent
\textbf{(2) Evidence Sufficiency Evaluation.}
To assess the sufficiency and reliability of retrieved evidence, we leverage the advanced reasoning capabilities of ChatGPT-4 through a set of structured evaluation prompts, as illustrated in Box~\hyperref[Box-c]{A.3}.
These prompts are designed to comprehensively evaluate source credibility, evidence relevance, and inter-evidence consistency, thereby ensuring a rigorous and systematic assessment process.
However, since ChatGPT operates as a black-box model without transparency or reproducibility, its direct outputs are unsuitable for deployment.
Instead, we treat ChatGPT’s evaluations as pseudo-labeled supervision signals to train an open-source backbone, Qwen2.5-14B-Instruct.
This design allows the model to acquire task-specific reasoning and evaluation capabilities while preserving interpretability and reproducibility.

\noindent
\textbf{(3) Instruction-Tuning.}
We construct training tuples $(X_t, E_t, C_t, C_r)$, where $X_t$ denotes the news item, $E_t$ represents the evidence retrieved by the agent, and $(C_t, C_r)$ correspond to the pseudo-labeled sufficiency label and its rationale generated by ChatGPT.
Using these tuples, we fine-tune Qwen2.5-14B-Instruct via LoRA (Low-Rank Adaptation of Large Language Models) ~\citep{55}, updating only low-rank adapter matrices while freezing the original model parameters.
This approach enables efficient adaptation to the evidence evaluation task, effectively integrating external pseudo supervision into an interpretable, open-source LLM framework.

\subsubsection{\textbf{Evaluation-Aware Evidence Expert.}}

When the evaluator determines that the retrieved evidence is sufficiently supportive, the evidence expert is activated to transform evidence assessment into actionable decision-making.
The evaluator transfers both the evidence $E_t$ and its reasoning rationale $R_t$, which describes the reliability and sufficiency of each evidence item, including justifications for trustworthy sources and concerns about potentially unreliable ones, to the expert for final judgment.

Unlike conventional detectors that rely solely on confidence scores obtained by assessing media credibility for evidence~\cite{60}, our expert explicitly models the dependency between the evidence and its evaluation rationale.
This design enables evaluation-aware reasoning, allowing the expert to adapt its decisions based on interpretable evidence-quality assessments grounded in a logical chain of reasoning.
As illustrated in Figure~\ref{fig:framework}, the expert comprises three components, detailed as follows.

\noindent
\textbf{1) Evidence Analyzer.} 
To capture the interaction between the evidence $E_t$ and its evaluation rationale $R_t$, we introduce an evidence analyzer formulated as:
\begin{equation}
\hat{C}_t = \mathrm{sigmoid(MLP(\mathbf{F_t}))}, \quad L_a = L_{ce}(\hat{C}_t, C_t),
\end{equation}
where $\mathbf{F_t}$ denotes the BERT-encoded representation of the concatenated text $\left[E_t;R_t\right]$; $\hat{C}_t$ is the predicted sufficiency label; and $L_{ce}$ is the cross-entropy loss.
This formulation allows the expert to internalize the evaluator’s reasoning process and adaptively calibrate its reliance on evidence, thereby improving robustness to unreliable or noisy external information.




\noindent
\textbf{2) News–Evidence Coordinator.}
To effectively leverage external evidence for detecting news authenticity, we design a news-evidence coordinator to model the semantic dependencies between news and evidence and to facilitate a more comprehensive understanding of their relationships.
The coordinator is implemented using a cross-attention mechanism, formulated as:
\begin{equation}
\mathrm{f_{cross}}(\mathbf{Q}, \mathbf{K}, \mathbf{V}) = \mathrm{softmax}\left(\frac{\mathbf{Q} \mathbf{K}^T}{\sqrt{d_k}}\right)\mathbf{V},
\end{equation}
\begin{equation}
\mathbf{v_{e\&x}} = \mathrm{f_{cross}}(\mathbf{F}_t, \mathbf{X}_t, \mathbf{X}_t).
\end{equation}
where $\mathbf{Q}$, $\mathbf{K}$, and $\mathbf{V}$ represent the query, key, and value matrices, respectively.

\noindent
\textbf{3) Prediction Classifier.}
A masked attention mechanism \citep{55} is applied to the news representation $\mathbf{X}_t$ to filter out irrelevant or noisy tokens, producing a refined representation $\mathbf{v_x}$.
Finally, a classifier integrates both $\mathbf{v_x}$ and the evidence-aware representation $\mathbf{v{e\&x}}$ to predict news authenticity:
\begin{equation}
L_{cls} = L_{ce}(\text{MLP}([\mathbf{v_x}; \mathbf{v_{e\&x}}]), y),
\end{equation}
where $[\cdot]$ denotes concatenation and $y \in \left\{0,1\right\}$ is the ground-truth label of news authenticity.
The overall loss for training the expert is expressed as:
\begin{equation}
L = \alpha L_a + L_{cls},
\end{equation}
where $\alpha$ is a weighting coefficient.




\subsection{Reasoning and Sentiment-Based Fallback}

\noindent
\textbf{Reasoning-Based Evaluator and Expert.}
\label{3.2}
When external evidence is insufficient, EASE discards the evidence and activates a reasoning-based strategy, which leverages the world knowledge and logical reasoning capabilities of LLMs to make judgments.
Similar to external evidence, the reasoning knowledge produced by LLMs can also be unreliable due to hallucinations.
To address this, the reasoning-based strategy mirrors the evidence-based framework by incorporating an evaluator for reasoning knowledge evaluation and a reasoning expert for decision-making.

Specifically, a reasoning agent first generates inference chains from the LLM for each news piece. The reasoning evaluator and expert are then trained using an instruction-tuning strategy similar to that employed in the evidence-based strategy. 
The prompts designed for the reasoning agent and evaluation are shown in Boxes~\hyperref[Box-a]{A.1} and ~\hyperref[Box-d]{A.4}.


\noindent
\textbf{Sentiment Expert.}
When both evidence and reasoning are deemed unreliable, EASE activates a fallback mechanism based on sentiment analysis.
As shown in Box~\hyperref[Box-b]{A.2}, the sentiment expert examines stylistic and emotional cues in the text, such as exaggerated, provocative, or biased expressions, to provide subjective signals that go beyond the factual content of the news.
It shares the same structural design as the evidence- and reasoning-based experts.

This cascaded architecture is trained in an end-to-end manner, forming a robust, flexible, and interpretable decision pipeline tailored to each news instance based on the availability and reliability of external information.

\section{RealTimeNews-25 Benchmark}

To facilitate the research of real-time fake news detection, we introduce a new RealTimeNews-25 benchmark that comprises recent news released in the past year.

\noindent \textbf{Data Collection.} We collected news articles from sources such as NBC News and BBC News, covering the period from June 2024 to September 2025. These articles span diverse domains, including politics, sports, and business. 
Moreover, most articles fall beyond the knowledge cutoffs of the LLMs used, which helps prevent potential data contamination in the model's internal knowledge.

\noindent \textbf{Data Synthesis Pipeline.} To generate instances of fake news, we utilize an LLM-based synthesis pipeline that modifies names, locations, and dates; interprets information out of context; alters writing styles; and inserts factual errors, such as using  ``America'' instead of ``Germany''. Each sample is manually verified against its true counterpart to ensure the presence of factual inaccuracies or logical inconsistencies. 

Importantly, to better simulate real-world real-time fake news detection, we \textbf{mask} the true source of each news item during evidence retrieval. This design
1) reflects practical scenarios where newly emerging news typically lack authoritative evidence, and
2) prevents information leakage by ensuring the model cannot rely on the news source itself for classification.
Under this setting, the model must retrieve and reason over alternative sources for supporting evidence.

\noindent \textbf{Comparison with Existing Benchmarks.}
As shown in Table \ref{tab:new_dataset}, existing benchmarks consist primarily of historical news released several years ago, whereas RealTimeNews-25 includes 3,487 more recent news articles.
To assess the degree of evidence scarcity across benchmarks, we randomly sample 100 news items from each dataset and manually determine whether any retrieved evidence is reliable based on human judgment.
We define the \textbf{evidence scarcity ratio} as the proportion of news items without reliable evidence to the total number of samples.
RealTimeNews-25 exhibits a scarcity ratio of 0.37, approximately twice that of other benchmarks.
Although the ratio of 0.37 is not particularly high, it reflects the inherent difficulty of collecting real-time news with reliable labels, while still offering valuable insights into fake news detection under evidence scarcity.

\section{Experiments}


\subsection{Experiment Setup}



\noindent
\textbf{Implementation Details.}
(1) \textbf{Agents.}
The agents for evidence retrieval, reasoning knowledge generation, and sentiment analysis use GPT-4o\footnote{\url{https://openai.com/}} for English datasets and Qwen2.5-VL\footnote{\url{https://bailian.console.aliyun.com/}} as the backbone
for Chinese datasets, respectively. Each agent is guided by task-specific prompts and interacts through the official APIs. The prompts used for reasoning knowledge and sentiment acquisition are provided in Appendix~\ref{Appendix-A}. 
The evidence retrieval process is limited to a maximum of 3 iterations.
(2) \textbf{Evaluators.} 
The evidence and reasoning evaluators are built on Qwen2.5-14B-Instruct\footnote{\url{https://huggingface.co/Qwen/Qwen2.5-14B-Instruct}}
 for Chinese datasets and Llama-3.1-8B-Instruct\footnote{\url{https://huggingface.co/meta-llama/Llama-3.1-8B-Instruct}}
 for English datasets.
All models are trained for two epochs using the AdamW optimizer~\citep{67} with LoRA fine-tuning~\citep{55}, a batch size of 4, and initial learning rates of $10^{-4}$ and $5\times10^{-5}$, for Chinese and English datasets, respectively. Training is performed on two NVIDIA A6000 GPUs.
(3) \textbf{Experts.}
We adopt BERT~\citep{65} as the text encoder in the expert module, using the bert-base-uncased''\footnote{\url{https://huggingface.co/google-bert/bert-base-uncased}} checkpoint for English datasets and the bert-base-chinese''\footnote{\url{https://huggingface.co/google-bert/bert-base-chinese}}
 checkpoint for Chinese datasets.
The maximum token length is set to 256, truncating longer sequences. Models are optimized with the Adam optimizer~\citep{66}, a learning rate of $2\times10^{-4}$ and a weight decay of $5\times10^{-5}$. The hidden dimension $d_k$ is 768. Early stopping with a patience of five epochs is applied to mitigate overfitting. The loss weighting factor $\alpha$ is set to 2 for Chinese datasets and 1 for English datasets.

\begin{table}[t]
\centering
\caption{Comparison of fake news detection benchmarks.}
\resizebox{0.5\textwidth}{!}{

\begin{tabular}{lccccc}
\toprule
\textbf{Datasets} & \textbf{Language} & \textbf{Period} & \textbf{\#Real} & \textbf{\#Fake}  & \textbf{Evidence Scarcity Ratio} \\ 
\midrule

Weibo \citep{57} & Chinese & 2020   & 4749 & 4779 & 0.21   \\ 
Weibo21 \citep{58} & Chinese & 2021   & 4640 & 4487 & 0.19   \\ 
GossipCop \citep{59} & English & 2017   & 10259 & 2581 & 0.15   \\
RealTimeNews-25 & English & 2024-2025  & 1963 & 1524 & 0.37  \\ \bottomrule
\end{tabular}
}
\label{tab:new_dataset}
\end{table}

\noindent\textbf{Evaluation Metrics.} We report accuracy (Acc.), macro F1 (macF1), and class-wise F1 scores for the fake and real classes (F1$_\mathrm{fake}$ and F1$_\mathrm{real}$).

\begin{table*}[ht]
    \centering  
        \caption{Performance comparison on the real-time fake news detection dataset RealTimeNews-25 and three historical datasets: Weibo, Weibo21, and GossipCop.
Models marked with $\dagger$ utilize external knowledge, while the others do not.
Models marked with $\ddagger$ indicate fact-checking approaches, whereas the remaining ones are fake news detection models.}
    \resizebox{\textwidth}{!}{
    \begin{tabular}{lcccc|cccccccccccc}
    \toprule
      \multicolumn{1}{c}{\multirow{2}{*}{\textbf{Model}}} & \multicolumn{4}{c|}{\textbf{RealTimeNews-25}} & \multicolumn{4}{c}{\textbf{Weibo}} & \multicolumn{4}{c}{\textbf{Weibo21}} & \multicolumn{4}{c}{\textbf{GossipCop}} \\
      \cmidrule(lr){2-5} \cmidrule(lr){6-9} \cmidrule(lr){10-13} \cmidrule(lr){14-17}
      & Acc. & macF1 & F1$_\mathrm{fake}$ & F1$_\mathrm{real}$ & Acc. & macF1 & F1$_\mathrm{fake}$ & F1$_\mathrm{real}$ & Acc. & macF1 & F1$_\mathrm{fake}$ & F1$_\mathrm{real}$ & Acc. & macF1 & F1$_\mathrm{fake}$ & F1$_\mathrm{real}$\\
    \midrule
        EANN~\cite{46} & 0.499 & 0.444 & 0.268 & 0.619 & 0.827 & 0.827 & 0.829 & 0.825 & 0.870 & 0.869 & 0.862 & 0.875 &  0.864 & 0.757 & 0.594 & 0.920 \\
         SAFE~\cite{47} & 0.544 & 0.502 & 0.356 & 0.647 & 0.762 & 0.761 & 0.774 & 0.748 & 0.905 & 0.896 & 0.901 & 0.890 & 0.838 & 0.769 & 0.643 & 0.895 \\
         CAFE~\cite{49} & 0.497 & 0.474 & 0.363 & 0.585 & 0.840 & 0.840 & 0.842 & 0.837 & 0.882 & 0.881 & 0.885 & 0.876 &  0.867 & 0.754 & 0.587 & 0.921\\
         BMR~\cite{50} & 0.571 & 0.558 & 0.481 & 0.635 & 0.918 & 0.909 & 0.914 & 0.904 & 0.929 & 0.926 & 0.927 & 0.925 &  0.895 & 0.813 & 0.691 & 0.936\\
         FND-CLIP~\cite{51} & 0.494 & 0.428 & 0.235 & 0.622 & 0.907 & 0.908 & 0.908 & 0.907 & 0.943 & 0.943 & 0.940 & 0.946 &  0.880 & 0.783 & 0.638 & 0.928\\
         MIMoE-FND~\cite{5} & 0.501 & 0.432 & 0.235 & 0.630 & 0.928 & 0.928 & 0.928 & 0.927 & 0.956 & 0.956 & 0.955 & 0.957 & 0.895 & 0.817 & 0.698 & 0.936\\ \midrule
         GET$^{\dagger}$~\cite{52} & - & - & - & - & 0.666 & 0.662 & - & - & 0.847 & 0.773 & - & - &  - & - & - & -\\
         SEE$^{\dagger}$~\cite{53} & - & - & - & - & 0.932 & 0.932 & - & - & 0.864 & 0.807 & - & -&  - & - & - & - \\
         ARG$^{\dagger}$~\cite{37} & 0.557 & 0.538 & 0.443 & 0.632 & 0.904 & 0.904 & 0.901 & 0.906 & 0.932 & 0.932 & 0.933 & 0.931 & 0.863 & 0.770 & 0.624 & 0.916\\
         Hiss$^{\dagger, \ddagger}$~\cite{10} & 0.621 & 0.619 & 0.590 & 0.646 & - & - & - & - & 0.652 & 0.643 & - & - & 0.798 & 0.659 & 0.432 & 0.887\\
         Veract Scan$^{\dagger, \ddagger}$~\cite{24} & 0.612 & 0.609 & 0.574 & 0.644 & - & - & - & - & - & - & - & - & - & - & - & - \\
         DEFAME$^{\dagger, \ddagger}$~\cite{14} & 0.698 & 0.694 & \textbf{0.729} & 0.658 & 0.831 & 0.829 & 0.811 & 0.847 & 0.819 & 0.819 & 0.825 & 0.813 & 0.822 & 0.682 & 0.471 & 0.893\\
         EASE$^{\dagger}$ (Ours) & \textbf{0.756} & \textbf{0.754} & 0.728 & \textbf{0.779}  & \textbf{0.933} & \textbf{0.933} & \textbf{0.933} & \textbf{0.932} & \textbf{0.962} & \textbf{0.962} & \textbf{0.962} & \textbf{0.961} &  \textbf{0.904} & \textbf{0.836} & \textbf{0.731} & \textbf{0.942}\\
     \bottomrule
    \end{tabular}
    }
    \label{mainresult}
    \vspace{-0.2cm}
\end{table*}

\subsection{Comparison with State-of-the-Art Methods}
We conduct experiments on three widely used fake news detection benchmarks, Weibo~\citep{57}, Weibo21~\citep{58}, and GossipCop~\citep{59}, to evaluate model performance on historical news with relatively sufficient external evidence.
In addition, we evaluate our newly introduced RealTimeNews-25 benchmark to assess model performance on more recent news under evidence scarcity.

As shown in Table~\ref{mainresult}, we systematically compare representative state-of-the-art (SOTA) models for fake news detection and fact-checking, which are grouped into two categories based on whether they incorporate external knowledge.  
When evidence is insufficient, fact-checking models such as Hiss~\citep{10}, Veract Scan~\citep{24}, and DEFAME~\citep{14} output ``Not Enough Information (NEI)'', which we treat as ``Fake'' when calculating detection accuracy.
Statistically, these models produce NEI labels for 49.6\%, 33.0\%, and 45.6\% of the news samples on RealTimeNews-25, respectively, reflecting the challenge posed by evidence scarcity in real-time scenarios.

\noindent
\textbf{Historical News Detection.}
EASE consistently outperforms all competing methods across all evaluation metrics on Weibo, Weibo21, and GossipCop, demonstrating its strong capability in detecting historical fake news where abundant external evidence is available.

\noindent
\textbf{Real-Time News Detection.}
Compared with the three historical datasets, RealTimeNews-25 presents a more challenging setting, resulting in a substantial decline in detection accuracy across existing SOTA methods due to their limited generalization ability to emerging events.
We observe that methods leveraging external knowledge (marked with $\dagger$ in Table~\ref{mainresult}) generally achieve higher performance than those without, suggesting that external information helps mitigate out-of-distribution issues and improve generalization to unseen news.
More importantly, EASE achieves an accuracy of 0.756, significantly outperforming comparable models. This improvement can be attributed to its carefully designed mechanisms that effectively evaluate and address insufficient external evidence and mitigate unreliable reasoning knowledge.

\begin{figure}
     \centering
     \begin{subfigure}[b]{0.237\textwidth}
         \centering
         \includegraphics[width=\textwidth, trim=0 0 0 0.01cm, clip]{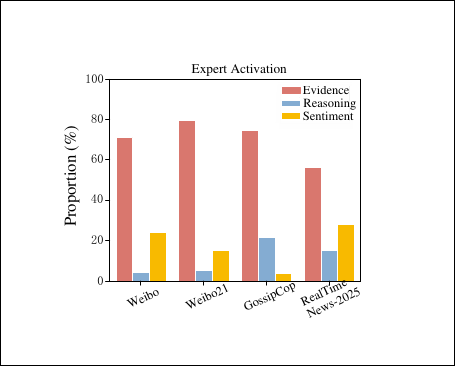}
         \caption{}
         \label{fig:activation}
     \end{subfigure}  
     \begin{subfigure}[b]{0.235\textwidth}
         \centering
         \includegraphics[width=\textwidth]{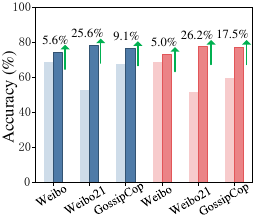}
         \caption{ }
         \label{fig:direct}
     \end{subfigure}
        \caption{(a) Activation ratios of different experts in EASE across datasets. (b) Evaluation accuracy of evaluators. \colorbox{mycolor1}{\rule{0pt}{0.8ex}\rule{0.8ex}{0pt}} \colorbox{mycolor2}{\rule{0pt}{0.8ex}\rule{0.8ex}{0pt}} represent the evidence evaluator before and after fine-tuning, respectively, while \colorbox{mycolor3}{\rule{0pt}{0.8ex}\rule{0.8ex}{0pt}} \colorbox{mycolor4}{\rule{0pt}{0.8ex}\rule{0.8ex}{0pt}} represent the reasoning evaluator before and after fine-tuning, respectively.}
        \label{fig:diff_dataset_expert}
          \vspace{-.15in}
\end{figure}

\subsection{Analysis of Expert Selection}
We analyze the ratios of experts selected for final decision-making in EASE across different datasets, as shown in Fig.~\ref{fig:diff_dataset_expert} (a).
Several insightful observations can be made:

1) The evidence expert is consistently activated most frequently across all datasets, followed by the sentiment expert and the reasoning expert, indicating that EASE primarily relies on retrieved evidence for decision-making.
2) The activation frequency of the evidence expert is noticeably lower on RealTimeNews-25 compared with the historical datasets, suggesting that real-time detection faces greater challenges in acquiring reliable supporting evidence.
3) Interestingly, even when the true source of the news is masked during evidence retrieval on RealTimeNews-25, the evidence expert remains the most active among all experts. This demonstrates that the evidence agent, through multi-source and multi-iteration retrieval, can still identify reliable or plausible evidence from alternative sources.


\begin{table*}[ht]
    \centering  
    \caption{Performance comparison of EASE against its variants with different expert modules ablated. The best results are highlighted in bold. Expert activation states are indicated by \textcolor{green}{\ding{51}} (activated) and \textcolor{red}{\ding{55}} (deactivated), with EE, RK, and SA representing the External Evidence, Reasoning Knowledge, and Sentiment Analysis experts, respectively.}
    \resizebox{\textwidth}{!}{
    \begin{tabular}{ccccccccccccccccccc}
    \toprule
      \multicolumn{3}{c}{\textbf{Expert}} & \multicolumn{4}{c}{\textbf{RealTimeNews-25}} & \multicolumn{4}{c}{\textbf{Weibo}} & \multicolumn{4}{c}{\textbf{Weibo21}} & \multicolumn{4}{c}{\textbf{GossipCop}} \\
      \cmidrule(lr){1-3} \cmidrule(lr){4-7} \cmidrule(lr){8-11} \cmidrule(lr){12-15} \cmidrule(lr){16-19}
  EE & RK & SA & Acc. & macF1 & F1$_\mathrm{fake}$ & F1$_\mathrm{real}$ & Acc. & macF1 & F1$_\mathrm{fake}$ & F1$_\mathrm{real}$ & Acc. & macF1 & F1$_\mathrm{fake}$ & F1$_\mathrm{real}$ & Acc. & macF1 & F1$_\mathrm{fake}$ & F1$_\mathrm{real}$\\
    \midrule
     \textcolor{green}{\ding{51}} &  \textcolor{red}{\ding{55}} & \textcolor{red}{\ding{55}} & 0.607 & 0.591 & 0.511 & 0.672 & 0.881 & 0.881 & 0.874 & 0.888 & 0.933 & 0.933 & 0.934 & 0.932 &  0.879 & 0.787 & 0.648 & 0.927\\
     \textcolor{red}{\ding{55}} &  \textcolor{green}{\ding{51}} & \textcolor{red}{\ding{55}} & 0.576 & 0.554 & 0.454 & 0.653 & 0.892 &  0.892 & 0.887 & 0.897 & 0.925 & 0.925 & 0.926 & 0.924 &  0.873 & 0.782 & 0.641 & 0.923\\
     \textcolor{red}{\ding{55}} &  \textcolor{red}{\ding{55}} & \textcolor{green}{\ding{51}} &  0.535 & 0.515 & 0.416 & 0.614 & 0.879 & 0.879  & 0.873 & 0.884 & 0.910 & 0.910 & 0.911 & 0.909 & 0.861 & 0.774 & 0.633 & 0.915\\
     \textcolor{green}{\ding{51}} &  \textcolor{green}{\ding{51}} & \textcolor{red}{\ding{55}} &  0.659 & 0.652 & 0.604 & 0.700 & 0.916 & 0.917 & 0.915 & 0.918 & 0.931 & 0.931 & 0.932 & 0.930 & 0.891 & 0.813 & 0.693 & 0.933\\
     \textcolor{green}{\ding{51}} &  \textcolor{red}{\ding{55}} & \textcolor{green}{\ding{51}} & 0.657 & 0.650 & 0.601 & 0.698 & 0.921 & 0.921 & 0.919 & 0.922 & 0.939 & 0.940 & 0.940 & 0.939 & 0.881 & 0.795 & 0.662 & 0.928\\
     \textcolor{red}{\ding{55}} & \textcolor{green}{\ding{51}} & \textcolor{green}{\ding{51}} & 0.601 & 0.585 & 0.504 & 0.667 & 0.898 & 0.897 & 0.892 & 0.901 & 0.940 & 0.940 & 0.941 & 0.938 &  0.866 & 0.767 & 0.616 & 0.919\\
     \textcolor{green}{\ding{51}} &  \textcolor{green}{\ding{51}} & \textcolor{green}{\ding{51}}  & \textbf{0.756} & \textbf{0.754} & \textbf{0.728} & \textbf{0.779} & \textbf{0.933} & \textbf{0.933} & \textbf{0.933} & \textbf{0.932} & \textbf{0.962} & \textbf{0.962} & \textbf{0.962} & \textbf{0.961} &  \textbf{0.904} & \textbf{0.836} & \textbf{0.731} & \textbf{0.942}\\
     \bottomrule
    \end{tabular}
    }
    \label{main_ablation}
\end{table*}

\begin{table}[!ht]
  \centering
  \caption{Effects of $L_a$ in different experts.}
  \scriptsize
  \setlength{\tabcolsep}{2pt}  
  \renewcommand{\arraystretch}{1.1}  
  \begin{tabularx}{\linewidth}{>{\centering\arraybackslash}p{1.4cm} *{4}{>{\centering\arraybackslash}X} *{4}{>{\centering\arraybackslash}X}}
    \toprule
    \multirow{2}{*}{\textbf{$L_a$}} & \multicolumn{2}{c}{\textbf{RealTimeNews-2025}} &
    \multicolumn{2}{c}{\textbf{Weibo}} & \multicolumn{2}{c}{\textbf{Weibo21}} & \multicolumn{2}{c}{\textbf{GossipCop}} \\
    \cmidrule(lr){2-3} \cmidrule(lr){4-5} \cmidrule(lr){6-7} \cmidrule(lr){8-9}
    & \multicolumn{1}{c}{Acc.} & \multicolumn{1}{c}{macF1} & \multicolumn{1}{c}{Acc.} & \multicolumn{1}{c}{macF1} & \multicolumn{1}{c}{Acc.} & \multicolumn{1}{c}{macF1} & \multicolumn{1}{c}{Acc.} & \multicolumn{1}{c}{macF1} \\
    \midrule
    \rowcolor[HTML]{E6E6E6} \multicolumn{9}{c}{\textbf{Evidence Expert}} \\
    \midrule
    \textcolor{red}{\ding{55}} & 0.575 & 0.570 & 0.875 & 0.875 & 0.915 & 0.916 & 0.853 & 0.764\\
    \textcolor{green}{\ding{51}} & 0.607 & 0.591 & 0.881 & 0.881 & 0.933 & 0.933 & 0.879 & 0.787  \\ 
    \midrule
    \rowcolor[HTML]{E6E6E6} \multicolumn{9}{c}{\textbf{Reasoning Expert}} \\
    \midrule
    \textcolor{red}{\ding{55}} & 0.556 & 0.549 & 0.850 & 0.850 & 0.890 & 0.891 & 0.828 & 0.739 \\
    \textcolor{green}{\ding{51}} & 0.576 & 0.554 & 0.892 & 0.892 & 0.904 & 0.903 & 0.873 & 0.782  \\
    \midrule
    \rowcolor[HTML]{E6E6E6} \multicolumn{9}{c}{\textbf{Sentiment Expert}} \\
    \midrule
    \textcolor{red}{\ding{55}} & 0.504 & 0.494 & 0.827 & 0.827 & 0.875 & 0.875 & 0.813 & 0.716 \\
    \textcolor{green}{\ding{51}} & 0.535 & 0.515 & 0.879 & 0.879 & 0.910 & 0.910 & 0.861 & 0.774 \\
    \bottomrule
  \end{tabularx}
  \label{tab:ablation_analyzer}
\end{table}





\subsection{Ablation Study}

\subsubsection{\textbf{Effect of Perspective-Specific Experts.}}

To evaluate the contribution of each expert in EASE, we conduct an ablation study by selectively removing one or two experts, as reported in Table~\ref{main_ablation}.

1) Single-expert removal.
When removing one expert, excluding the evidence expert leads to the largest performance drop in accuracy on RealTimeNews-25, Weibo, and GossipCop, underscoring the crucial role of external evidence in EASE.
In contrast, on Weibo21, the greatest decline occurs when the sentiment expert is removed. Further analysis reveals that Weibo21 contains a large number of user-generated posts written in emotional and colloquial language, where external evidence and structured reasoning are less informative. This highlights the importance of sentiment analysis in detecting community-driven rumors.

2) Dual-expert removal.
When two experts are removed and the model relies on a single expert, performance degrades significantly across all datasets. This indicates that depending solely on a single perspective without reliability evaluation severely limits the model’s decision-making ability.
Overall, these results demonstrate the necessity of incorporating multiple complementary experts together with reliability-aware evaluation to achieve robust fake news detection.

\subsubsection{\textbf{Analysis of Evaluators}}

The LLM-based evaluators in EASE are fine-tuned with pseudo labels generated by ChatGPT using an instruction-tuning strategy.
To assess their effectiveness, we investigate three key questions:
1) Are the pseudo labels reliable?
2) Does fine-tuning improve the evaluators’ assessment accuracy?
3) Does fine-tuning further enhance the overall detection performance?
The following experiments address these questions in sequence.

\noindent \textbf{1) Are the pseudo labels reliable?}
We conduct a human evaluation to examine the consistency between ChatGPT-generated pseudo labels and human judgments.
Ten annotators assessed the reliability of evidence for 100 randomly sampled news–evidence pairs per dataset, resulting in 400 samples in total.
The evaluation criteria followed those used for prompting ChatGPT, as described in Box~\ref{Box-C} (Appendix).
A similar process was applied to reasoning knowledge.
Results show that \textbf{93\%} of the samples received consistent reliability labels from both annotators and ChatGPT, indicating a strong alignment between human and model evaluations.

\noindent \textbf{2) Does fine-tuning improve the evaluators’ assessment accuracy?}
We further evaluate the evaluators’ reliability before and after fine-tuning through another round of human evaluation.
Five annotators rated 400 samples (100 per dataset), following the same procedure as in the pseudo-label evaluation.
A similar setup was used for the reasoning evaluator, substituting reasoning knowledge for external evidence.
As shown in Figure~\ref{fig:diff_dataset_expert} (b), both fine-tuned evidence and reasoning evaluators substantially improve accuracy, confirming that fine-tuning enhances their alignment with task-specific evaluation criteria and mitigates the noise in direct LLM outputs.

\noindent \textbf{3) Does fine-tuning enhance the overall detection accuracy?}
We compare the performance of EASE using evaluators with and without fine-tuning.
As shown in Table~\ref{tab:ablation_finetune}, fine-tuned evaluators consistently yield higher accuracy  across all datasets.
This improvement validates that enhancing evaluator reliability directly strengthens evaluation-aware decision-making in subsequent fake news detection.

\begin{table*}[htbp]
\small
\caption{Open-world real-time case study of EASE. News samples were collected in real time beyond datasets. EASE activates the evidence-, reasoning-, and sentiment-based experts for the three cases, respectively.
}
\begin{tabular}{lll}
\toprule
\multicolumn{2}{p{12cm}}{{\bf News-1}: UN personnel have been detained by Houthi forces, and Guterres strongly condemns the act. Guterres expressed deep concern about the safety of UN personnel in Yemen and reiterated his call for the Houthi authorities to immediately and unconditionally release all detained individuals. \color{blue}{[Label: REAL]}} & \makecell[c]{\multirowcell{15}{\vspace{2.5cm}\includegraphics[width=.25\textwidth]{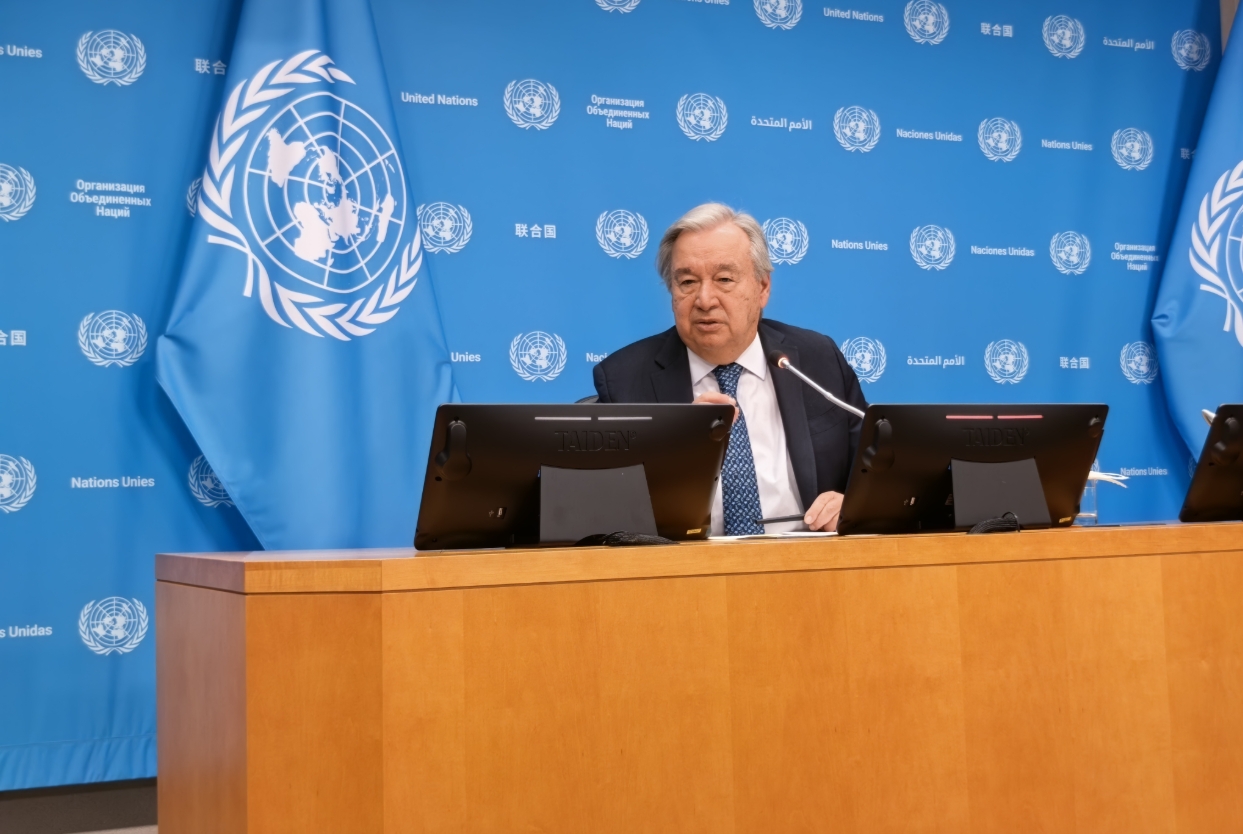}}}  \\
\cmidrule{1-2}
\multicolumn{1}{p{12cm}}{{\bf Evidence}:  UN personnel have been detained by Houthi forces in Yemen, and this action has been condemned by UN Secretary-General António Guterres. \par
{\bf Evidence Evaluator}: The evidence is reliable due to its authoritative source, comprehensive information, and consistent internal logic.\par
{\bf Evidence Expert}: \color{blue}{REAL} \color{green}{\ding{51}}}
\\
\midrule
\midrule
\multicolumn{2}{p{12cm}}{{\bf News-2}: On October 6th, a march occurred in the capital of Venezuela, Alaskan, with participation from social movement organizations, the People's Front, community groups, and many citizens. A collective protest expressing opposition to external threats. \color{blue}{[Label: FAKE]}} & \makecell[c]{\multirowcell{15}{\vspace{1.8cm}\includegraphics[width=.25\textwidth]{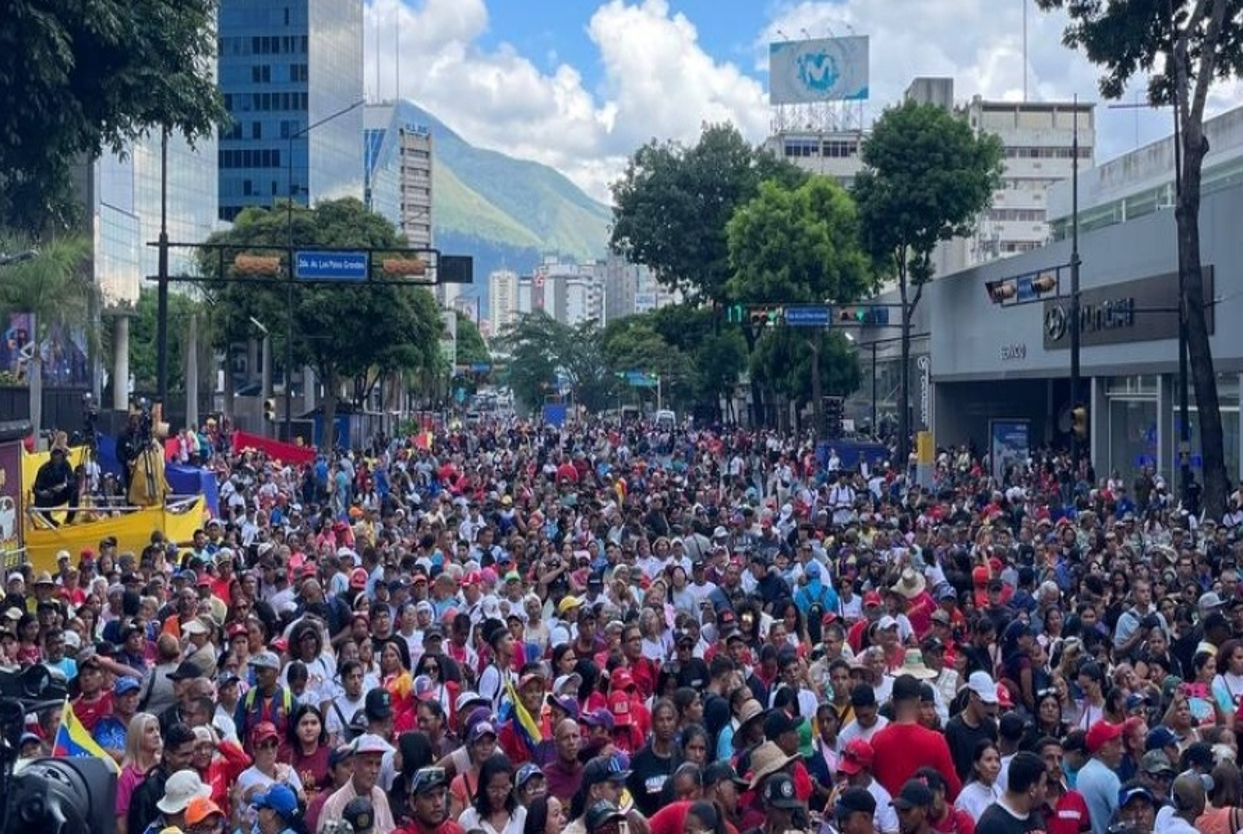}}}  \\
\cmidrule{1-2}
\multicolumn{2}{p{12cm}}{{\bf Evidence}:  The news lacks sufficient evidence and cannot be verified based on the current information.\par
{\bf Evidence Evaluator}: The evidence is unreliable due to insufficient evidence.\par
{\bf Reasoning Knowledge}: The capital of Venezuela is Caracas, not "Alaskan." Alaska is a state in the United States. \par
{\bf Reasoning Evaluator}: I have enough information to judge whether the news is real or not.\par
{\bf Reasoning Expert}: \color{blue}{FAKE} \color{green}{\ding{51}}}
\\
\midrule
\midrule
\multicolumn{2}{p{12cm}}{{\bf News-3}: Barcelona was hit by a terrifying power outage! ⚡ The entire city was plunged into complete darkness for HOURS. I was in bed when the lights suddenly went out — total silence. At first, I thought it was a minor glitch, but it quickly became clear: this was BIG. Where has the government been? What have they been doing? \color{blue}{[Label: FAKE]}} & \makecell[c]{\multirowcell{15}{\vspace{-0.2cm}\includegraphics[width=.25\textwidth]{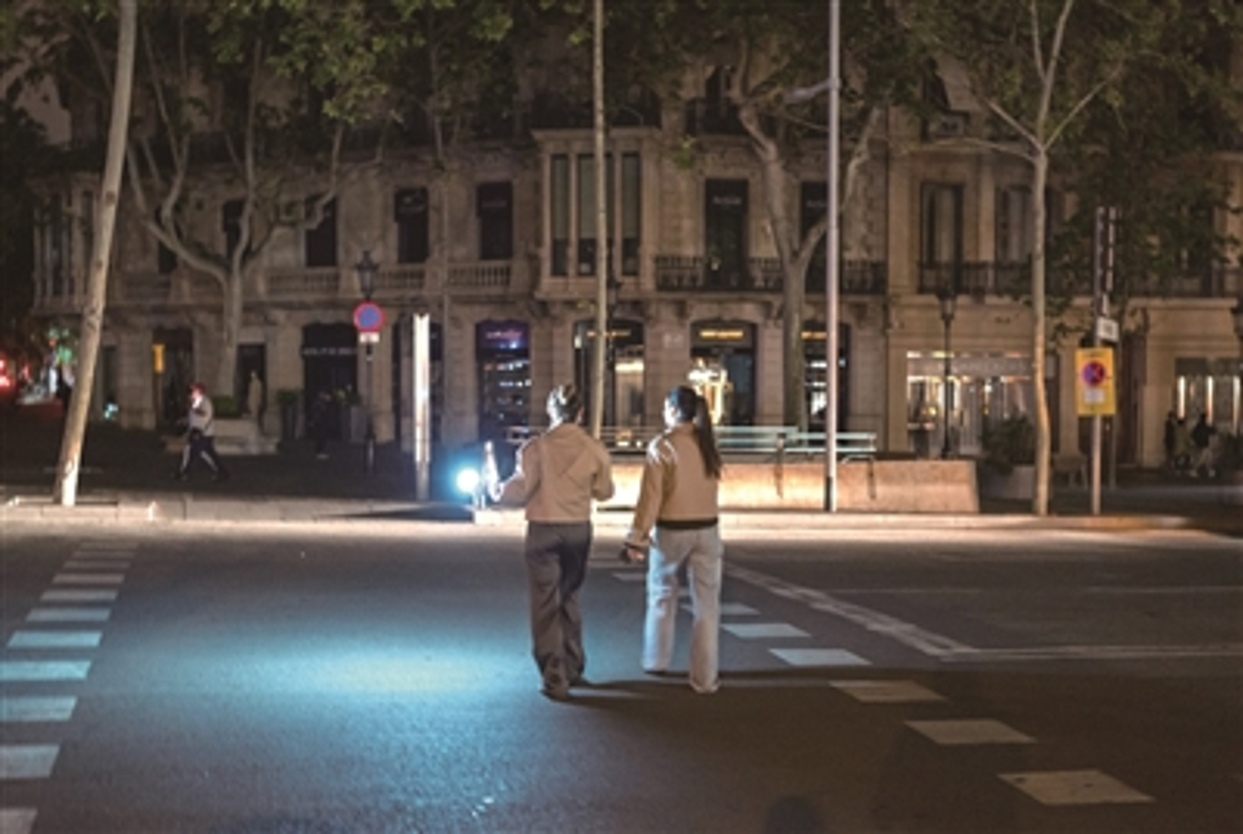}}}  \\
\cmidrule{1-2}
\multicolumn{1}{p{12cm}}{{\bf Evidence}:  Evidence retrieval indicates that a large-scale power outage occurred in Barcelona in April 2025, but there is no indication of a similar event recently.\par
{\bf Evidence Evaluator:} The evidence is unreliable because it contains contradictory information.\par
{\bf Reasoning Knowledge:} A large-scale power outage lasting several hours due to system overload or equipment aging is possible in major cities like Barcelona. However, without the specific time of the incident, I cannot make an accurate judgment.\par
{\bf Reasoning Evaluator:} I don't have enough information to judge whether the news is real or not. \par
{\bf Sentiment Analysis:} The use of terms like "terrifying," "total silence," and "I was in bed" creates a tense atmosphere. This emotionally driven content is often used in misinformation to evoke an emotional response rather than convey factual information.\par
{\bf Sentiment Expert}: \color{blue}{FAKE} \color{green}{\ding{51}}}

\\
\bottomrule
\end{tabular}
\label{tab:cases}
\end{table*}

\subsubsection{\textbf{Analysis of Losses}}

In addition to the primary classification loss $L_{cls}$ for news authenticity detection, EASE introduces an auxiliary loss $L_a$ for the analyzer within each expert module.
As shown in Table~\ref{tab:ablation_analyzer}, removing $L_a$ leads to a noticeable drop in detection accuracy across all datasets.
This result confirms that $L_a$ enables experts to internalize both the knowledge and its evaluation from the evaluators, thereby enhancing their evaluation-aware decision-making ability rather than merely relying on externally provided assessments.

\begin{table}[!ht]
  \centering
  \caption{Effects of fine-tuning evaluators.}
  \scriptsize
  \setlength{\tabcolsep}{2pt}  
  \renewcommand{\arraystretch}{1.1}  
  \begin{tabularx}{\linewidth}{>{\centering\arraybackslash}p{1.2cm} *{4}{>{\centering\arraybackslash}X}}
    \toprule
    \textbf{Fine-Tuning} 
    & Acc.
    & macF1 
    &  F1$_\mathrm{fake}$ 
    & F1$_\mathrm{real}$ \\
     \midrule
    \rowcolor[HTML]{E6E6E6} \multicolumn{5}{c}{\textbf{Weibo}} \\
    \midrule
    \centering\textcolor{red}{\ding{55}} & 0.893 & 0.894 & 0.895 & 892  \\
    \centering\textcolor{green}{\ding{51}} & 0.933	& 0.933	& 0.933	& 0.932 \\
    \midrule
    \rowcolor[HTML]{E6E6E6} \multicolumn{5}{c}{\textbf{Weibo21}} \\
    \midrule
    \centering\textcolor{red}{\ding{55}} & 0.927 & 0.927 & 0.924 & 0.929  \\
    \centering\textcolor{green}{\ding{51}} & 0.962	& 0.962	& 0.962	& 0.961 \\
    \midrule
    \rowcolor[HTML]{E6E6E6} \multicolumn{5}{c}{\textbf{GossipCop}} \\
    \midrule
    \centering\textcolor{red}{\ding{55}} & 0.874 & 0.778 & 0.631 & 0.925   \\
    \centering\textcolor{green}{\ding{51}} & 0.904	& 0.836	& 0.731	& 0.942 \\
    \midrule
    \rowcolor[HTML]{E6E6E6} \multicolumn{5}{c}{\textbf{RealTimeNews-25}} \\
    \midrule
    \centering\textcolor{red}{\ding{55}} & 0.669 & 0.664 & 0.622 & 0.706   \\
    \centering\textcolor{green}{\ding{51}} & 0.756	& 0.754	& 0.728	& 0.779 \\
    \bottomrule
  \end{tabularx}
    \label{tab:ablation_finetune}
\end{table}

\subsection{Open-World Real-Time Case Study}

To further evaluate the real-world applicability of our method beyond benchmark datasets, we conduct an open-world case study.
As shown in Table~\ref{tab:cases}, the selected news items were released in October 2025, only a few days before paper submission, and thus exhibit a realistic challenge of evidence scarcity.
To prevent information leakage, we masked the original sources from which the news was collected.

In the first example, EASE successfully retrieves sufficient and relevant alternative evidence, enabling the evidence expert to make an accurate verification.
The second example contains a factual inconsistency, where “Alaskan” is incorrectly mentioned as the capital of Venezuela instead of “Caracas.”
In this case, when external evidence is insufficient, the reasoning expert leverages commonsense knowledge to correctly identify the news as fake.
The third example features an emotionally charged post describing a power outage in “Barcclona,” lacking temporal specifics and containing conflicting evidence.
Here, logical reasoning offers limited support due to ambiguous factual grounding, but the sentiment expert effectively recognizes exaggerated and emotionally biased language, leading EASE to correctly classify it as fake.

These open-world cases demonstrate the strong robustness and generalizability of EASE in real-time fake news detection, even under conditions of scarce or unreliable supporting evidence.


\section{Limitations}


1) Dataset limitations. Although we aim to study the evidence scarcity issue by collecting recent news from the past year, the degree of scarcity in RealTimeNews-25 (Table \ref{tab:new_dataset}) is not significantly high, even though it surpasses existing benchmarks. This reflects the inherent difficulty of obtaining real-time news with reliable labels. Nevertheless, our open-world real-time case study (Table \ref{tab:cases}) demonstrates that EASE generalizes well to emerging news events.
2) Method limitations. While we employ sentiment analysis as the final fallback mechanism, objectively written news lacking strong emotional cues can mislead the sentiment agent and expert, as shown in the failure case analysis in the Appendix. Future work will explore additional decision-making perspectives to improve robustness under such conditions.

\section{Conclusion}


In this work, we systematically investigate the practical challenge of evidence scarcity in real-time fake news detection.
We propose EASE, a sequential, multi-perspective evaluation and decision-making framework designed to explicitly assess evidential sufficiency and handle cases with insufficient evidence.
To support evaluation in real-world settings, we also introduce RealTimeNews-25, a new benchmark for real-time fake news detection.
Extensive experiments on RealTimeNews-25, together with three public benchmarks, demonstrate the superior performance of EASE on historical news and its strong generalization ability to emerging real-time events.

\bibliographystyle{ACM-Reference-Format}
\bibliography{sample-base}

\appendix

\section{Prompts for Knowledge Acquisition and Evaluation}\label{Appendix-A}

\begin{figure}[htb]
    \centering
    \includegraphics[width=\linewidth,height=0.5\textwidth,trim=10 10 10 10mm]{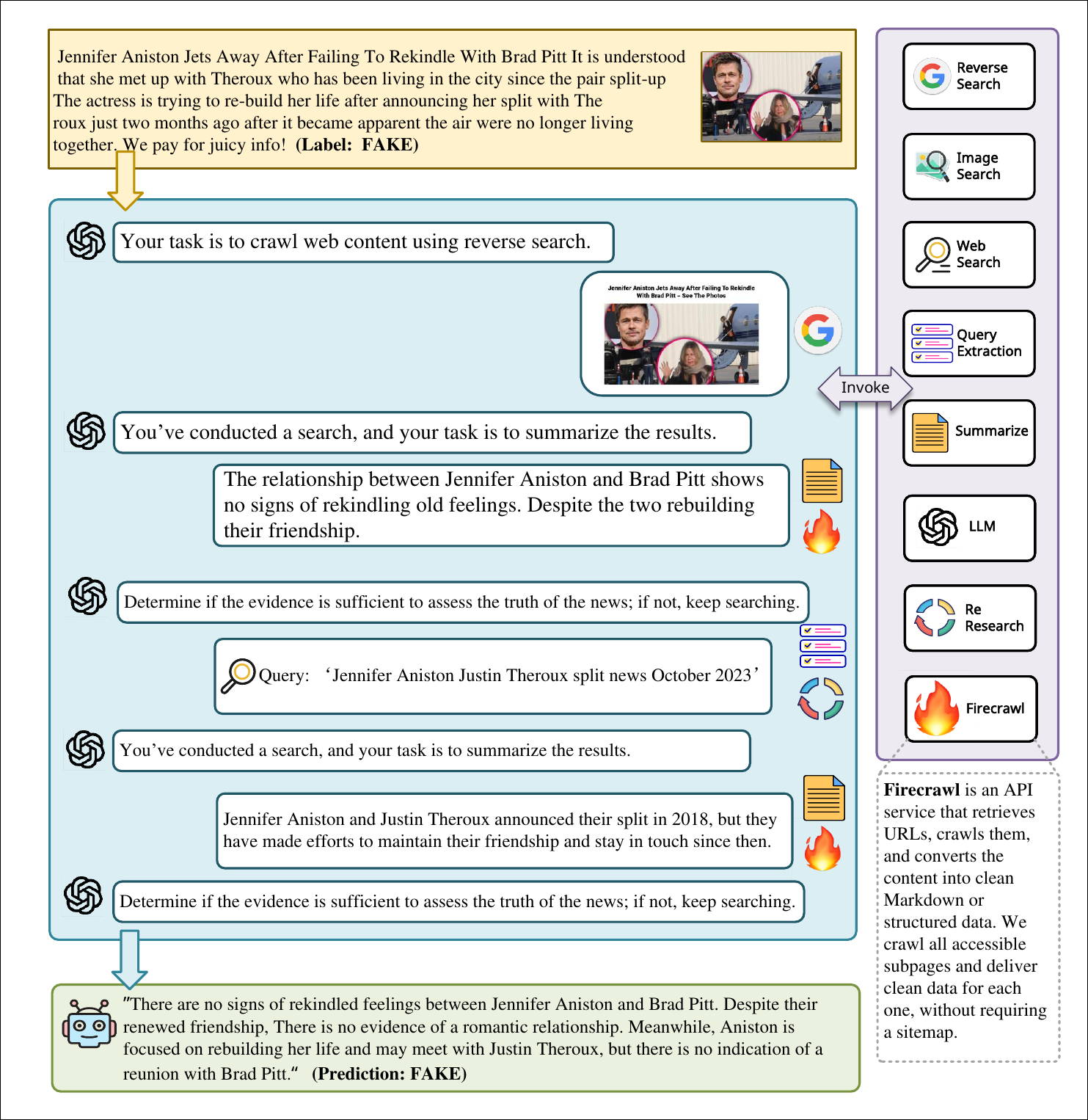}
    \caption{Prompting process of the evidence agent.}
    \label{fig:agent_prompt}
\end{figure}

Fig.~\ref{fig:agent_prompt} and Boxes~A.1–A.2 present the structured prompts designed for the evidence, reasoning, and sentiment agents to acquire perspective-specific knowledge.
Boxes~A.3–A.4 illustrate the prompts used by the evidence and reasoning evaluators to assess the reliability of the acquired knowledge.


\begin{numberedappendixbox}{Reasoning Knowledge Generation}\label{Box-a}
As a logical reasoning expert, please analyze the truthfulness of the following news content:\\
\textbf{News:} [news]\\
Analysis Criteria (Logic and Common Sense):\\
- Analyze from a common sense perspective, such as the political systems of various countries, living habits of animals, and other common knowledge\\
- Whether the cause-effect relationships follow logical patterns\\
- Whether this is physically possible according to scientific laws\\
- Whether any numbers or statistics are realistically feasible\\
Output Format:\\
Analysis: [2-3 sentence logical assessment]\\
Prediction: true/fake\\
Important: Keep the entire output under 190 tokens.\\
\end{numberedappendixbox}

\begin{numberedappendixbox}{Sentiment Analysis}\label{Box-b}
As a content analysis expert, please assess the truthfulness of the following news content by examining its emotional tone and sensationalism:\\
\textbf{News:} [news]\\
Analysis Criteria (Emotional and Sensational Language):\\
- Identify whether the language contains inflammatory or provocative elements\\
- Evaluate if the tone is excessively subjective or biased\\
- Detect the presence of exaggerated or alarmist statements\\
- Assess if emotional manipulation is being used to influence readers\\
Output Format:\\
Analysis: [2-3 sentence assessment focusing on emotional tone and sensationalism]\\
Prediction: true/fake\\
Important: Keep the entire output under 190 tokens.\\
\end{numberedappendixbox}

\begin{numberedappendixbox}{Prompt Design for Evidence Evaluator}
\label{Box-c}
\textbf{News:} [news]\\
\textbf{Related news evidence from search results:} [evidence]\\
Please evaluate the reliability of external evidence for the news story by applying the following three criteria:\\
1. Internal Consistency: Are the different pieces of evidence consistent with each other?\\
2. Information Sufficiency: Does the evidence provide sufficient details to support the news story?\\
3. Source Credibility: Are the sources of the evidence authoritative and trustworthy?\\
Judgment:\\
Sufficient: All dimensions are strong or there are only minor issues.\\
Insufficient: There are significant problems with the evidence.\\
Output:\\
Evidence [Sufficient/Insufficient], because [Short, specific reasons.].\\
\end{numberedappendixbox}

\begin{numberedappendixbox}{Prompt Design for Reasoning Evaluator}
\label{Box-d}
Evaluate the reliability of reasoning knowledge for news.\\
\textbf{News:} [news]\\
\textbf{Related reasoning knowledge from LLMs:} [reasoning knowledge]\\
Please evaluate the reliability of reasoning knowledge for the news by applying the following criteria:\\
1. \textbf{Confidence}: Do you have enough confidence in the logical reasoning knowledge to judge the truthfulness of the news? If not, consider the reasoning knowledge unreliable.\\
2. \textbf{Plausibility}: Does the reasoning follow logically from established facts and align with common sense? Are there any logical fallacies or implausible assumptions?\\
Judgment:\\
Reliable: All dimensions are strong, or there are only minor issues.\\
Unreliable: There are significant problems with the reasoning knowledge.\\
Output:\\
Evidence [Reliable/Unreliable], because [Short, specific reasons].\\
\end{numberedappendixbox}



\section{Inference Time Analysis}

We measure the inference time on 400 randomly sampled instances from the RealTimeNews-25 dataset.
As shown in Table~\ref{tab:inferene_time}, we compare EASE with three state-of-the-art (SOTA) models that achieve the most competitive detection performance: Hiss~\citep{10}, Veract Scan~\citep{24}, and DEFAME~\citep{14}.
The results reveal a clear trade-off between accuracy and inference time. Hiss and Veract Scan offer faster inference but at the cost of lower accuracy. In contrast, EASE not only substantially surpasses the strongest baseline, DEFAME, in detection accuracy but also achieves faster inference speed.

We further provide a detailed runtime breakdown of EASE in Table~\ref{tab:runtime_breakdown}.
The primary time cost arises from \textbf{evidence retrieval}, \textbf{reasoning generation}, and \textbf{sentiment analysis}, among which evidence retrieval is the most time-consuming. This is mainly due to the multi-round query mechanism designed to enhance the accuracy and relevance of retrieved evidence.
Since GPT models are accessed exclusively via OpenAI’s API, most computational workloads are executed externally. Specifically, evidence acquisition, reasoning generation, and sentiment analysis are performed through API calls to large language models. The internal components are executed on a dual-socket server equipped with two Intel Xeon Gold 6526Y CPUs, totaling 32 physical cores and 64 logical threads.

In contrast, the evaluator and expert modules introduce small overhead to the total inference time.



\begin{table}[h]
\centering
\caption{Comparison of inference time among competitive models.}
\begin{tabular}{lc}
\hline
 \textbf{Method} & \textbf{Inferene Time (min:sec)}  \\
\hline
         DEFAME~\cite{14} & 3:03 \\
         Hiss~\cite{10}  & 0:41  \\
         Veract\;Scan~\cite{24}  & 0:48 \\ 
         EASE & 1:23 $\sim$ 1:50  \\
\hline
\end{tabular}
\label{tab:inferene_time}
\end{table}

\begin{table}[h]
\centering
\caption{Inference time breakdown of EASE.}
\label{tab:runtime_breakdown}
\begin{tabular}{lc}
\toprule
\textbf{Component} & \textbf{Time (min:sec)} \\
\midrule
Evidence Retrieval & 1:23 \\
Evidence Evaluator & 0:05 \\
Evidence Expert & <0:01 \\
Reasoning Generation & 0:08 \\
Reasoning Evaluator & 0:05 \\
Reasoning Expert & <0:01 \\
Sentiment Analysis & 0:07 \\
Sentiment Expert & <0:01 \\
\midrule
Total (minimal\textasciitilde maximum) & 1:23\textasciitilde 1:50 \\
\bottomrule
\end{tabular}
\end{table}

\section{Illustrative Examples from RealTimeNews-25}

Box G presents a representative data sample from RealTimeNews-25, which includes rich contents, such as the news content, ground-truth label, external evidence, reasoning knowledge, and sentiment analysis. It also features initial assessments generated by LLMs from multiple perspectives, along with metadata that indicates whether the news was modified, its source, and reliability annotations for both the evidence and reasoning components. 

Table~\ref{tab:cases_realtimenews} presents examples of fake news generation on RealTimeNews-25, created by modifying original news articles collected from authoritative sources.

\begin{tcolorbox}[
    colback=gray!10!white,
    colframe=black,
    title=G. Data in RealTimeNews-25,
    fonttitle=\bfseries\small,
    fontupper=\small,before skip=5pt,after skip=5pt
]\label{Box-C}
\{\\
\hspace*{0.5cm}\textbf{ID:} 1 \\
\hspace*{0.5cm}\textbf{Content:} Death of Slim Shady: The controversial legacy of \\
\hspace*{0.5cm}Eminem's peroxide-blond alter ego \\
\hspace*{0.5cm}\textbf{Published:} 2024-06-01 \\
\hspace*{0.5cm}\textbf{Label:} real \\
\hspace*{0.5cm}\textbf{Evidence:} In his new album "The Death of Slim Shady (Coup  \\
\hspace*{0.5cm}De Grâce)", Eminem examines the controversial legacy... \\
\hspace*{0.5cm}\textbf{Evidence Prediction:} real \\
\hspace*{0.5cm}\textbf{Reasoning:} The title uses the provocative term "Death" and \\
\hspace*{0.5cm}frames the topic around a "controversial legacy"... \\
\hspace*{0.5cm}\textbf{Reasoning Prediction:} real \\
\hspace*{0.5cm}\textbf{Sentiment:} The title uses metaphorical language \\
\hspace*{0.5cm}common in music journalism... \\
\hspace*{0.5cm}\textbf{Sentiment Prediction:} real \\
\hspace*{0.5cm}\textbf{Modification Type:} origin \\
\hspace*{0.5cm}\textbf{Sources:} BBC \\
\hspace*{0.5cm}\textbf{Evidence Reliable:} 1 \\
\hspace*{0.5cm}\textbf{Reasoning Reliable:} 1\\
\},...
\end{tcolorbox}

\begin{table*}[h]
\small
\caption{Examples of data collection in the RealTimeNews-25 dataset. ``Modified News'' denotes fake news generated by altering the original news content indicated in red.}
\begin{tabular}{lll}
\toprule
\multicolumn{2}{p{12cm}}{{\bf News-1}: On March 25, local time, German Federal President Steinmeier will present the certificate of dismissal to the current Chancellor Scholz and his cabinet members. Scholz and his cabinet will continue to perform their duties until a new federal government is established.} & \makecell[c]{\multirowcell{15}{\vspace{3cm}\includegraphics[width=.2\textwidth]{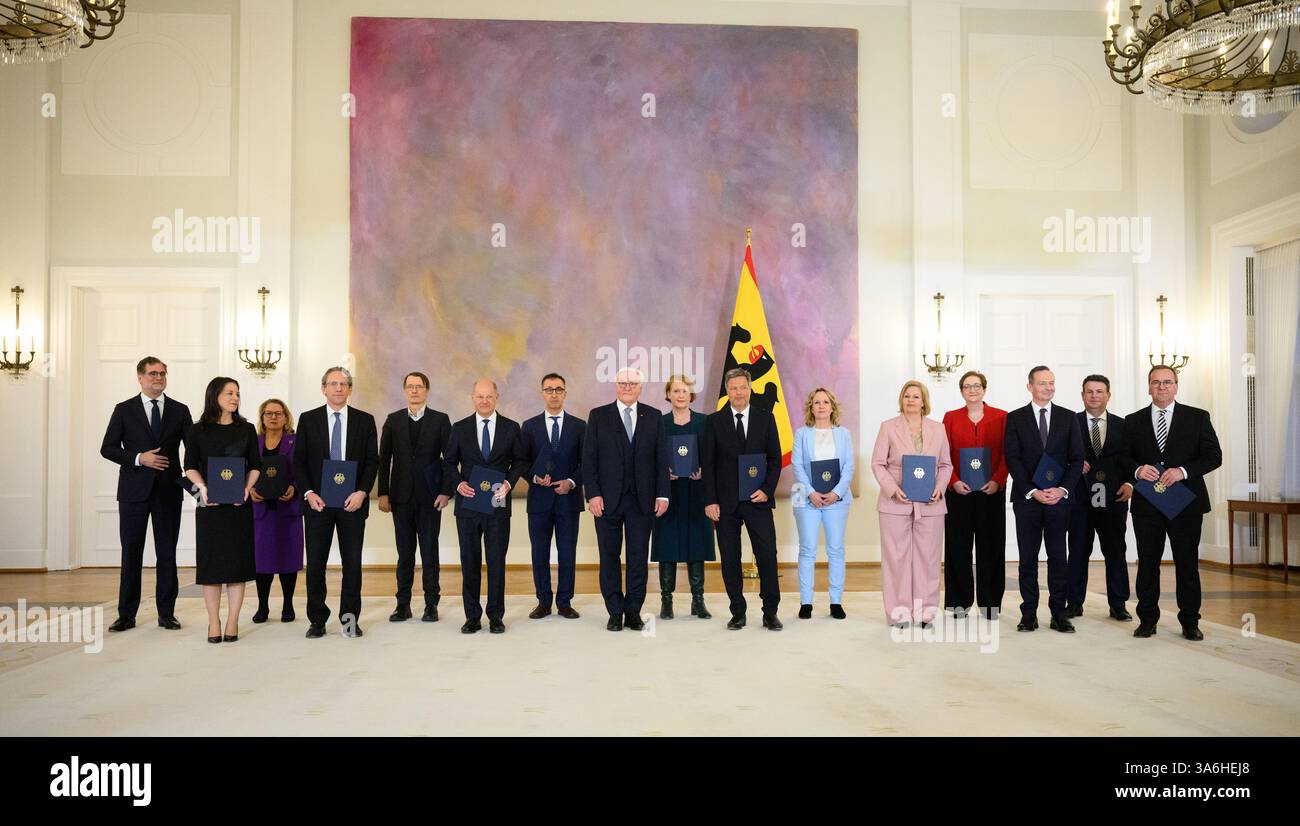}}}  \\
\cmidrule{1-2}
\multicolumn{1}{p{12cm}}{{\bf Modified News}: \textcolor{red}{On April 15}, local time, \textcolor{red}{French President Macron} will present the certificate of dismissal to the current Prime Minister Dupont and his cabinet members. Dupont and his cabinet will continue to perform their duties until a new national government is formed.}
\\
\midrule
\midrule
\multicolumn{2}{p{12cm}}{{\bf News-2}: As the Israel Defense Forces (IDF) conducted large-scale operations in the West Bank, destroying over 20 buildings in the Jenin refugee camp on February 2, Palestinian President Mahmoud Abbas called for an emergency meeting of the United Nations Security Council. He urged the United States to intervene in what he described as an Israeli “act of aggression.”} & \makecell[c]{\multirowcell{15}{\vspace{2cm}\includegraphics[width=.2\textwidth]{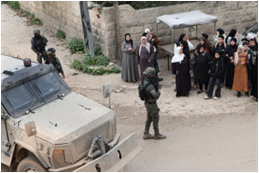}}}  \\
\cmidrule{1-2}
\multicolumn{2}{p{12cm}}{{\bf Modified News}:  \textcolor{red}{The Egyptian Armed Forces} continue large-scale operations in the Sinai Peninsula, and on \textcolor{red}{March 10}, they destroyed over 30 buildings in the Rafah refugee camp. \textcolor{red}{The President of Egypt, Sisi}, calls for an emergency meeting of the UN Security Council, urging the United States to intervene in Egypt's 'aggressive actions.'}
\\
\midrule
\midrule
\multicolumn{2}{p{12cm}}{{\bf News-3}: The Chinese national table tennis team's list of participants for the 2025 Doha World Table Tennis Championships has been announced, led by Wang Chuqin, Sun Yingsha, Lin Shidong and others.} & \makecell[c]{\multirowcell{15}{\vspace{5cm}\includegraphics[width=.2\textwidth]{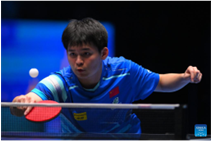}}}  \\
\cmidrule{1-2}
\multicolumn{1}{p{12cm}}{{\bf Modified News}: Retain the original news. \par
 \rule{0pt}{1.2cm}
}
\\
\bottomrule
\end{tabular}
\label{tab:cases_realtimenews}
\end{table*}

\begin{table*}[htb]
\small
\caption{Failure cases of EASE.}
\begin{tabular}{lll}
\toprule
\multicolumn{2}{p{12cm}}{{\bf News-1}: Climate change does not exacerbate the dangers posed by hurricanes. The World Meteorological Organization emphasizes that all naturally occurring climate events now take place against the backdrop of human-induced climate change. Driven by global warming, climate change intensifies extreme weather events and disrupts seasonal rainfall and temperature patterns. \textcolor{blue}{[Label: FAKE]}} & \makecell[c]{\multirowcell{15}{\vspace{3cm}\includegraphics[width=.25\textwidth]{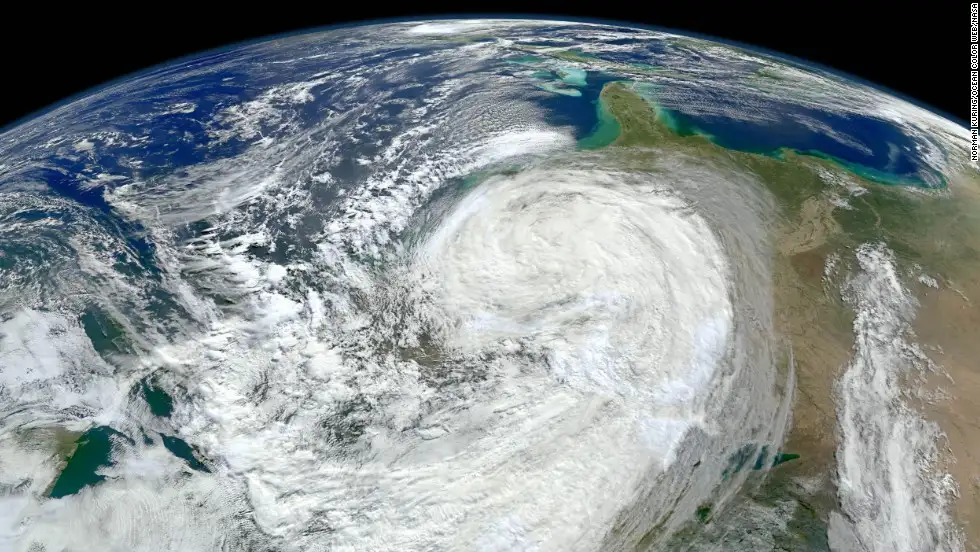}}}  \\
\cmidrule{1-2}
\multicolumn{1}{p{12cm}}{
{\bf External Evidence}: The claim that climate change exacerbates hurricane dangers is supported by multiple sources. The World Meteorological Organization reports an increase in tropical cyclone clusters in the North Atlantic due to global warming. However, some scientists caution against blaming Hurricane Sandy solely on climate change, noting that it is not yet possible to definitively link any single storm to global warming, and the connection is not as clear as some argue.\par
{\bf Evidence Evaluator}: The evidence contains contradictory information.\par
{\bf Reasoning Knowledge}: Climate change has not yet been definitively linked to an increase in the global frequency of hurricanes, but it has significantly intensified their strength, rainfall, and associated risks. Therefore, disaster prevention and mitigation strategies must place greater emphasis on the extreme impacts that hurricanes can bring.\par
{\bf Reasoning Evaluator}: The reasoning knowledge contains inconsistent statements, thus it is unreliable.\par
{\bf Sentiment Analysis:} The language is factual and neutral, citing an authoritative organization without inflammatory or provocative elements. The tone is objective, stating scientific observations without exaggerated or alarmist statements or any apparent emotional manipulation.\par
{\bf Sentiment Expert:} \textcolor{blue}{REAL} \textcolor{red}{\ding{55}}
}
\\
\bottomrule
\end{tabular}
\label{tab:errors}
\end{table*}

\section{Failure Case}

Table~\ref{tab:errors} presents a failure case of EASE.
In this example, the news claim that “Climate change does not exacerbate the dangers posed by hurricane” leads the agent to retrieve relevant but contradictory evidence. The reasoning knowledge states that although climate change is not conclusively linked to increased hurricane frequency, it intensifies storm strength, precipitation, and related risks. However, this reasoning is evaluated as unreliable due to internal inconsistencies. Moreover, the sentiment analyzer fails to capture emotional shifts or inflammatory cues within the neutrally toned official report. As a result, EASE fails to correctly assess this case.
Nevertheless, the presence of refuting evidence in external sources suggests the potential for developing more fine-grained evaluation mechanisms to better reconcile conflicting information and improve model robustness.

\end{document}